\pdfoutput=1

\documentclass[11pt]{article}

\usepackage[final]{acl}

\usepackage{times}
\usepackage{latexsym}

\usepackage[T1]{fontenc}

\usepackage[utf8]{inputenc}

\usepackage{microtype}

\usepackage{inconsolata}

\usepackage{graphicx}
\usepackage{url}
\usepackage{booktabs}
\usepackage{xspace}
\usepackage{placeins}
\usepackage[most]{tcolorbox}
\newcommand{\mistral}{{Mistral-7B-Instruct}\xspace}
\newcommand{\mistralev}{{Mistral-7B-Instruct with external knowledge}\xspace}
\newcommand{\commandr}{{Command R+}\xspace}
\newcommand{\commandrev}{{Command R+ with external knowledge}\xspace}

\title{Dynamic Knowledge Integration for Evidence-Driven Counter-Argument Generation with Large Language Models}

 \author{Anar Yeginbergen \and Maite Oronoz \and Rodrigo Agerri \\
         HiTZ Center - Ixa, University of the Basque Country UPV/EHU\\
         \texttt{\{anar.yeginbergen,maite.oronoz,rodrigo.agerri\}@ehu.eus}
         }

\begin{document}
\maketitle
\begin{abstract}

This paper investigates the role of dynamic external knowledge integration in improving counter-argument generation using Large Language Models (LLMs). While LLMs have shown promise in argumentative tasks, their tendency to generate lengthy, potentially non-factual responses highlights the need for more controlled and evidence-based approaches. We introduce a reconstructed and manually curated dataset of argument and counter-argument pairs specifically designed to balance argumentative complexity with evaluative feasibility. We also propose a new LLM-as-a-Judge evaluation methodology that shows a stronger correlation with human judgments compared to traditional reference-based metrics. Our experimental results demonstrate that integrating dynamic external knowledge from the web significantly improves the quality of generated counter-arguments, particularly in terms of relatedness, persuasiveness, and factuality. The findings suggest that combining LLMs with real-time external knowledge retrieval offers a promising direction for developing more effective and reliable counter-argumentation systems.
Data and code are publicly available.\footnote{\url{https://github.com/anaryegen/counter-argument-generation}}
\end{abstract}

\section{Introduction}

Argumentation in Natural Language Processing (NLP) is becoming an increasingly active area of research, driven by the natural human tendency to express disagreement with claims or viewpoints expressed by individuals during information exchanges. In fact, it is becoming an indispensable tool in many application domains such as public policy, law, medicine, and education \cite{stab-gurevych-2017-parsing,eger-etal-2018-cross,garcia-ferrero-etal-2024-medmt5,yeginbergen-etal-2024-argument,sviridova-etal-2024-casimedicos}.

\begin{figure}[ht]
\includegraphics[width=\linewidth]{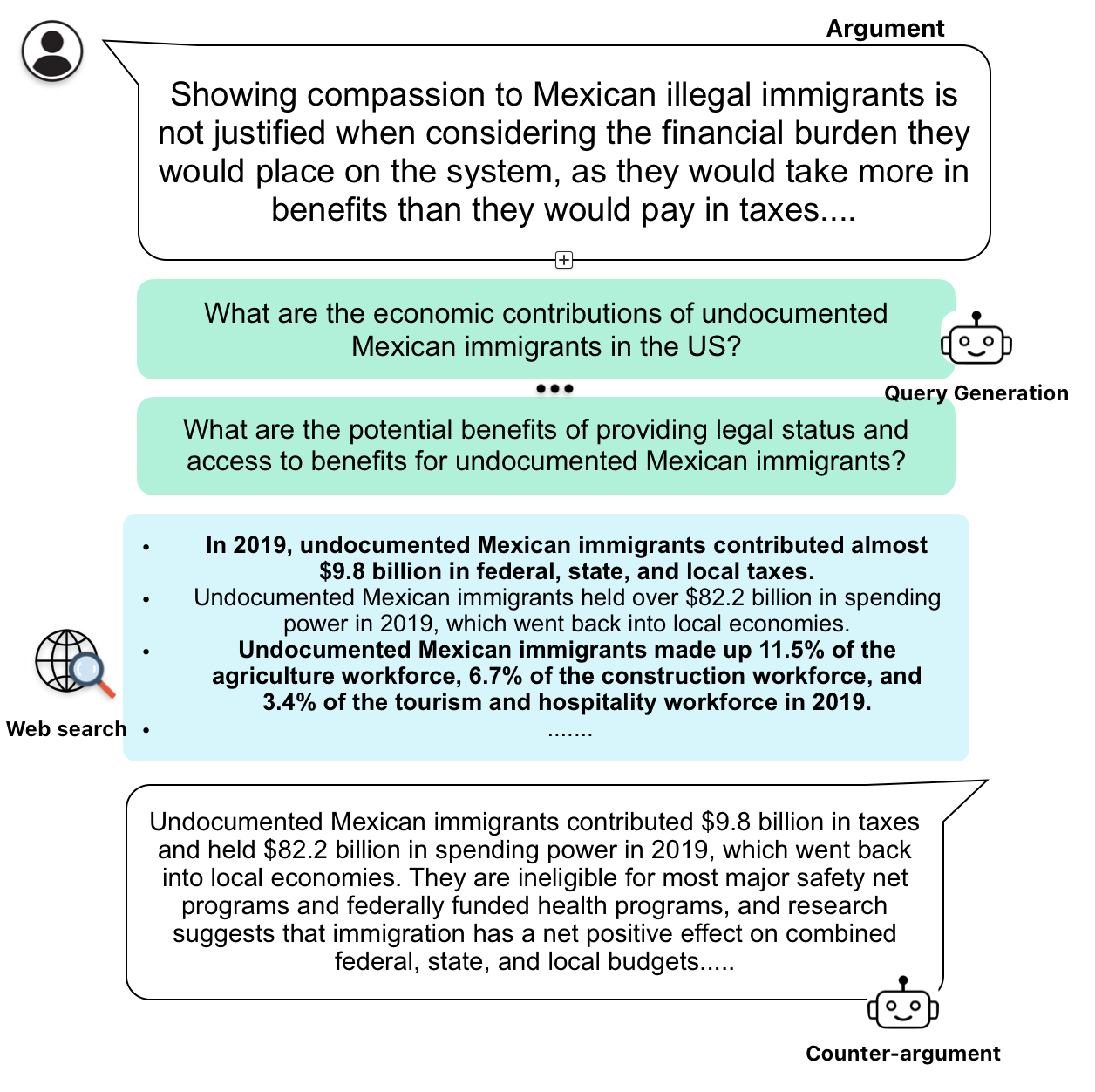}
\centering
\caption{Our approach to counter-argument generation integrating dynamic external knowledge.}
\label{fig:intro_fig}
\end{figure}

It is possible to distinguish two main research lines in Argumentation in NLP: (i) argument mining, which involves analyzing unstructured texts to automatically identify and extract argumentative elements and their relations \cite{cabrio2018five,stab-gurevych-2017-parsing,yeginbergen-etal-2024-argument}; (ii) argument generation, which focuses on generating argumentative texts using external knowledge sources \cite{hua-etal-2019-argument-generation,schiller-etal-2021-aspect} or summarizing key argumentative points \cite{roush2020debatesum,syed-etal-2021-generating,chen-etal-2024-exploring-potential}. 

This paper contributes to the second line of research by investigating whether dynamic integration of external knowledge helps LLMs to improve counter-argument generation. Counter-argument generation seeks to develop an effective framework for presenting alternative perspectives to an argument while ensuring the correctness of the message and incorporating factual evidence \cite{wachsmuth-etal-2017-argumentation,wachsmuth-etal-2018-retrieval}. LLMs have shown promising potential to deal with debates or any disagreements by solely relying on the parametric knowledge encoded in the model \cite{chen-etal-2024-exploring-potential,alshomary-wachsmuth-2023-conclusion}. Moreover, most of the LLMs have security safeguards to avoid harmful interactions by any means \cite{bai2022traininghelpfulharmlessassistant,NEURIPS2022_b1efde53,touvron-etal-2023-llama}. While maintaining harmlessness is recommendable, it is also important to be reasonable, persuasive, and grounded with a factual basis when arguing \cite{khandebating}.

Previous work on counter-argument generation has focused on specific aspects of the argument \cite{schiller-etal-2021-aspect, saha-srihari-2023-argu}, integrating different external news sources \cite{hua-etal-2019-argument-generation}, or by attacking different evidence of the initial argument \cite{jo-etal-2020-detecting,chen-etal-2024-exploring-potential}. However, no previous work has considered integrating dynamic knowledge to inform counter-argument generation. Furthermore, we argue that existing datasets for counter-argument generation consist of excessively long \cite{hua-wang-2018-neural,hua-etal-2019-argument-generation,chen-etal-2024-exploring-potential} or extremely short \cite{schiller-etal-2021-aspect,lin-etal-2023-argue} arguments, which makes it difficult to accurately evaluate their quality. Thus, although modern LLMs tend to generate rather lengthy essay-style responses that may be highly persuasive, they lack coherence, logic, and factual evidence, restricting them to short single sentences is insufficient to study the complexity and desired pragmatic characteristics of a good counter-argument \cite{hinton2023persuasive, ji2023survey, carrasco2024large, verma2024auditing}. Finally, the manual evaluation of counter-argument quality is a challenging, time-consuming, and highly subjective task that requires knowledge of the subject matter and credible evidence to support or refute it \cite{wachsmuth-etal-2017-argumentation, wang-etal-2017-winning, hua-etal-2019-argument-generation}. In this sense, we believe that traditionally used \cite{alshomary2021counter,chen-etal-2024-exploring-potential} reference-based automatic metrics such as BLEU, METEOR, or BERTScore fail to accurately capture the nuanced relationship between the generated counter-argument and the human judgment.

In order to address these issues, we propose using retrieved external knowledge from dynamic sources without any limits to particular outlets or databases, along with LLMs, to generate efficient, concise, evidence-based counter-arguments. Furthermore, we present a new curated dataset of concise and structured human-generated argument and counter-argument pairs in which the length of the counter-arguments is enough to study the main argumentative aspects while facilitating manual and automatic evaluation. While this paper employs both manual and automated evaluation methods, we introduce a novel automated evaluation approach using LLMs-as-a-Judge, designed to optimize correlation with human assessments.

Using the new dataset and dynamically retrieved external evidence, this work aims to answer the following research questions (RQ):
\begin{itemize}
    \item \textbf{RQ1}: Does the integration of dynamic external knowledge into LLMs help to generate better counter-argumentation?
    \item \textbf{RQ2:} Which automatic evaluation method correlates better with human judgments? 
    \item \textbf{RQ3:} To what extent do LLMs use retrieved external evidence in producing counter-arguments?
\end{itemize}

Figure \ref{fig:intro_fig} illustrates our proposed framework. First, we automatically generate queries that challenge the main points of the argument or claim for more fact-based information and feed those queries to the web search. Next, related evidence is retrieved, and, lastly, the claim and the retrieved evidence are provided as context to the LLM to generate a counter-argument. The main contributions of our paper are the following:

\begin{enumerate}
    \item We publicly share a new dataset of manually curated arguments and counter-arguments.
    \item A new method to dynamically integrate external knowledge retrieved from the web in LLM-based counter-argument generation.
    \item Experimental results show that the generation of counter-arguments with LLMs is improved through the integration of dynamic external knowledge, with factual evidence demonstrating a particularly significant impact on pragmatic aspects, including relatedness, persuasiveness, and factuality.
    \item Our automatic evaluation based on LLM-as-a-Judge reveals a higher correlation with human judgments compared with reference-based metrics such as BLEU, METEOR, or BERTscore.
\end{enumerate}

\section{Related Work}\label{sec:related-work}

Prior research in efficient and persuasive argument generation has approached the problem from various perspectives, focusing on different aspects of arguments. For instance, \citet{jo-etal-2020-detecting} investigated the ability of machine learning approaches to detect attackable sentences in the arguments, and they concluded that automatic approaches are more stable in this task than human annotators, depending on the subjectivity, topic, and tone of the argument. \citet{alshomary2021counter} focused on generating counter-arguments by pinpointing and challenging weak premises. \citet{schiller-etal-2021-aspect} produces arguments by specifying the desired aspect and stance in a sentence-level setting via Conditional Transformer Language Model (CTRL) \citep{keskar2019ctrl}. Similarly, \citet{saha-srihari-2023-argu} proposed a method to control both the topic and stance of an argument while enriching it with factual evidence using an encoder-decoder language model.
\citet{alshomary-wachsmuth-2023-conclusion} propose a counter-argument generation based on the stance of the conclusion of the argument. \citet{lin-etal-2023-argue} employed large language models for sentence-level counter-argument generation, implementing a Chain-of-Thought methodology. \citet{hu2023americano} introduces an agent-based chain-of-thought (CoT) method for generating arguments by decomposing the claim and sequentially processing the argument for the final output. Finally, \citet{chen-etal-2024-exploring-potential} evaluated LLMs in several argument mining and generation tasks, showing the potential of LLMs for this particular task.


The use of external sources has demonstrated its effectiveness in the generation of alternative perspectives. \citet{wachsmuth-etal-2018-retrieval} analyzed the question of retrieving the best counter-arguments when no prior knowledge is available. \citet{hua-wang-2018-neural} introduced a framework that incorporated information retrieval, but their approach was limited to using the Wikipedia database as the external source. However, Wikipedia primarily contains static factual information, which may not align with the dynamic nature of arguments. To address this limitation, \citet{stab-etal-2018-argumentext} expanded the scope by indexing all documents from the Common Crawl database for argument retrieval. \citet{hua-etal-2019-argument-generation} proposed an enhanced framework that leverages databases from news outlets alongside Wikipedia to retrieve evidence and improve the quality of the generated counter-arguments.

All these efforts mainly refer to static databases as external sources, meaning that all the documents containing the evidence of the argument in question should be parsed in advance. Moreover, previous argument generation with external knowledge was proposed before the LLMs entered the race. In our work, we believe that we should test the capabilities of LLMs and that, in our ever-changing dynamic world, it is not efficient to rely on a pre-defined set of documents as an external source for factual and persuasive counter-argument generation. Instead, we propose to integrate knowledge from the whole internet as a source for finding factual evidence to generate better counter-argumentation.

\section{Data}\label{sec:data}

In order to perform our experiments and to guarantee a balance in the length of input argument and output counter-argument in the data, we constructed a new corpus for the evaluation of counter-argument generation. Previous work often focused on either (too short) sentence-level \citep{lin-etal-2023-argue, schiller-etal-2021-aspect} or (too long) paragraph-level \citep{hua-etal-2019-argument-generation, hua-wang-2018-neural} generation.

Given the capabilities of modern LLMs, it was essential to specify clear input and output data to ensure accurate, robust, and fair comparisons during evaluation. Thus, our objective is to create a dataset of argument/counter-argument pairs that would meet specific criteria. 
We argue that (i) generating single-sentence claim-based counter-arguments is insufficient to accurately assess the quality of counter-arguments produced by LLMs and (ii), counter-arguments that are too long make it extremely difficult to properly evaluate the argumentative quality of the generated text.
Our analysis revealed that Large Language Models (LLMs), when not given explicit length constraints, tend to generate verbose, essay-like responses that frequently deviate from true argumentative form, lack substantiating evidence, and demonstrate poor coherence with the original argument while overemphasizing persuasive elements. To address these limitations, we propose generating counter-arguments with a maximum length of three sentences, focusing on conciseness, factual content, and direct alignment with the input argument.

With this aim in mind, we constructed a new dataset of argument and counter-argument pairs using the \texttt{CANDELA} corpus as a basis \cite{hua-etal-2019-argument-generation}. The corpus consists of debates and discussions on various controversial topics from the \texttt{r/ChangeMyView}\footnote{\url{https://www.reddit.com/r/changemyview}} subreddit. The corpus is centred around real-world online debates where users post their opinions and evidence for any controversial topic and expect other users to provide reasoning for an alternative perspective.

\begin{table}[ht]
  \centering
  \begin{tabular}{l|cc}
    \hline
                    & \# sentence & \# words \\
    \hline
    arguments && \\
    \hline
    Original        & 16   & 372  \\
    Intermediate      & 3    & 83   \\
    Final           & 3    & 61   \\
    \hline 
    counter-arguments && \\
    \hline
    Original       & 30 & 921  \\
    Intermediate     & 5  & 165   \\
    Final          & 3  & 72   \\
    \hline
  \end{tabular}
  \caption{Average number of sentences and words in arguments and counter-arguments in the original, summarized (intermediate), and final versions of the data.}
  \label{tab:data_averages}
\end{table}


\texttt{CANDELA} is available in a format split by sentences, tokenized, and lowercased, making the reconstruction of the corpus necessary. To address this, we employed LLMs to convert the data back into a coherent, human-readable format.

Once the data was fully reconstructed in a human-readable format, we summarized all arguments and counter-arguments. To avoid any bias, we choose a language model different from those used in our experimentation, namely, Llama-3.1-70B-Instruct \cite{dubey2024llama}. The generated summaries were then \textit{manually verified} against the original data and re-summarized when required to ensure semantic and pragmatic correctness.


While the corpus includes data from real-world interactions and reflects arguments from natural exchanges of information, we found that not all topics were equally suitable to evaluate whether external knowledge benefits counter-argument generation. Specifically, topics lacking a scientific or factual foundation, namely, expressed on deeply subjective topics, often trigger LLMs to produce generic, safe responses due to their safety guardrails. Moreover, the external knowledge of the topics that lack ground-truth factual backing tends to be ambiguous and/or biased. An example of such arguments can be found in Appendix \ref{appendix: rejected_examples}.

To mitigate potential quality issues, we implemented a manual filtering process for the corpus, retaining only those arguments that demonstrated both high quality and direct relevance to the subject matter, specifically selecting instances where the incorporation of factual information would meaningfully contribute to argument generation.

Finally, we further manually refined the summaries to follow a structured argumentative format, emphasizing components such as the main claim, supporting evidence, and examples where applicable (see examples in Appendix \ref{appendix: more_data_examples}). This process resulted in a dataset of 150 high-quality 3-sentence paragraph-level argument and counter-argument pairs. 

Examples of the final version of the corpus used in the experiments can be seen in Appendices \ref{appendix: data_example} and \ref{appendix: more_data_examples}. The average distribution of sentences and tokens in the data is shown in Table \ref{tab:data_averages}.

\section{Experimental setup}\label{sec:experimental-setup}

In this section, we describe the methodology for generating counter-arguments with dynamic knowledge integration for evidence-driven counter-argument generation. 
Preliminary experiments revealed that relying on static pre-defined document sets (such as Wikipedia) for factual evidence retrieval often yields incomplete information regarding specific argument topics, resulting in incoherent and unreliable evidence. We determined that external sources must contain topic-specific content, opinions, and observations directly relevant to the events described in the argument, particularly since claims may reference recent events that post-date the last update of pre-parsed sources. Consequently, our experimental design incorporates \emph{dynamic} web-based data as the primary information source to address these limitations.


External knowledge is obtained through the web search tool provided by the Cohere API\footnote{\url{https://cohere.com/}}. The process involved automatically generating five queries (averaging 67 words each) designed to challenge the validity and veracity of the original argument's claims and premises by questioning key points that required factual substantiation. These queries were sequentially submitted to the web search engine, and the retrieved results, averaging 5,496 words in length, were incorporated as contextual information in the final prompt presented to the language model. The model then generated counter-arguments based on both the original argument and the retrieved contextual information.


To assess the role of external knowledge in counter-argument generation with LLMs, we performed a comparative analysis using two system configurations: one incorporating external information and another relying solely on the model's parametric knowledge. The model in the latter configuration received only the original argument as input and was tasked with generating counter-arguments using its internal knowledge base exclusively. This experimental design enabled us to evaluate whether LLMs display better performance when provided with external evidence for counter-argument generation, as measured by comparing the quality of outputs between the two configurations.

The experimentation is performed using two LLMs with strong results on text generation tasks in two different configurations: (i) the LLM using only the claim as a prompt to rely on its parametric knowledge and (ii) the LLM with dynamic external knowledge. This results in the following four models:
\begin{itemize}
    \item \textbf{\commandr}: Command-R+, a 104B parameter model from \citet{cohere_for_ai_2024}. 
    \item \textbf{\commandrev}: Command-R+ 104B with external evidence retrieved using Cohere's API for web search.
    \item \textbf{\mistral}: Mistral-7B-Instruct-v0.3 \cite{jiang2023mistral}.
    \item \textbf{\mistralev}: Mistral-7B-Instruct-v0.3 with external evidence retrieved via Cohere's API for web search.
\end{itemize}

Importantly, all experiments are conducted in an inferential setting under default hyperparameters to assess the real capabilities of LLMs in counter-argument generation. This setup ensures a fair and robust evaluation of their performance in generating meaningful, well-reasoned, and factual counter-arguments. A list of all the prompts used at each generation step is illustrated in Appendix \ref{appendix: prompt_examples}.


\subsection{Evaluation}

Following \citet{hua-etal-2019-argument-generation}, we assess the quality of the generated counter-arguments using a point-wise 3-point Likert scale across five key dimensions: Opposition, Relatedness, Specificity, Factuality, and Persuasiveness.

It should be noted that the evaluation of generated counter-arguments is inherently subjective. While human evaluation is the gold standard, it is time-intensive, costly, and prone to individual biases \cite{wachsmuth-etal-2017-argumentation,hua-etal-2019-argument-generation,chen-etal-2024-exploring-potential}. To address these limitations, along with human evaluation, we will also provide two types of automatic evaluation. First, using reference-based metrics such as BLEU, METEOR, or BERTscore previously used in counter-argumentation generation \cite{alshomary2021counter,chen-etal-2024-exploring-potential}. Second, we propose to use the LLM-as-a-Judge approach. Leveraging the consistency, scalability, and efficiency of LLMs, this method enables rapid and reproducible scoring across the five dimensions. To the best of our knowledge, we are the first to propose an LLM-based evaluation for counter-argument generation.

Taking the gold standard counter-argument dataset built in Section \ref{sec:data}, we generate a counter-argument for each claim using the four models listed above. The five counter-arguments (gold reference included) are then evaluated by human annotators and LLM-as-a-Judge annotators. More specifically, they are asked to assess each dimension by categorizing the example as \textit{unsatisfactory (score: 1)}, \textit{moderately satisfactory (score: 2)}, or \textit{highly satisfactory (score: 3)} for each of the five dimensions listed above. Finally, reference-based metrics were computed by comparing the generated counter-arguments with the gold reference.

We will also compute the correlation between both automatic evaluation methods and human judgments, as establishing strong alignment is crucial for ensuring valid comparisons and reducing reliance on human evaluators, ultimately leading to a more robust and practical evaluation framework.

\noindent \textbf{Human evaluation.} We recruited human evaluators through the crowdsourcing platform Prolific\footnote{\url{https://www.prolific.com/}} to assess the quality of a sample of 75 generated counter-arguments across the five predefined dimensions. We performed two rounds of evaluations of three participants to ensure that the obtained manual judgments were of high quality. We set test questions in the questionnaire to determine whether the evaluations were performed fairly. Subsequently, we selected the four participants who correctly passed the test. All evaluators were compensated above the minimum rate recommended by the platform. An example of instructions provided for manual evaluation can be found in Appendix \ref{appendix: google_form_example}.

\noindent \textbf{LLM-as-a-Judge.} We employ two state-of-the-art open LLM-as-a-Judge models, namely, Prometheus \cite{kim2024prometheus} and JudgeLM \cite{zhu2023judgelm}, and one proprietary model, Claude 3.5 Sonnet\footnote{\url{https://www.anthropic.com/claude/sonnet}}. The models are prompted with the same set of instructions for evaluation as those used for human annotators.

\noindent \textbf{Reference-based evaluation.} Following the previous evaluations on counter-argument generation \cite{hua-etal-2019-argument-generation, lin-etal-2023-argue,chen-etal-2024-exploring-potential} we evaluate using reference-based overlap and similarity metrics, such as BLEU \citep{papineni2002bleu}, ROUGE \citep{lin-2004-rouge}, METEOR \citep{banerjee2005meteor}, and BERTScore \citep{zhang2019bertscore}. BLEU and ROUGE compare the overlap between the counter-argument and the claim, whereas METEOR also considers synonyms, paraphrases, and word stems. Finally, BERTScore takes into account the semantic context and meaning of the text, going beyond surface-level word matching.

\noindent \textbf{Ranking}. Based on the evaluation scores per dimension, we obtain rankings of each of the 5 candidate counter-arguments (4 automatically generated and the gold reference) by summing the scores for each counter-argument, as provided by the human and LLM-as-a-Judge evaluators. Formally, let the evaluation process involve \( n \) counter-arguments, \( m \) dimensions, and \( e \) evaluators. Let \( Score_{i,j,d} \) represent the score assigned by evaluator \( j \) for counter-argument \( i \) on dimension \( d \).

The total score for counter-argument \( i \), summed across all dimensions for evaluator \( j \), is given by:

\[
T_{i,j,d} = \sum_{d} Score_{i,j,d}
\]


The final ranking for counter-argument \( i \), combining scores across all \( m \) dimensions for each evaluator \( j \), is calculated as:

\[
R_{i,j} = Rank(\sum_{d=1}^{m} T_{i,j,d}) 
\]

Where:
\begin{itemize}
    \item \( T_{i,j, d} \) is the total aggregated score for counter-argument \( i \) for dimension \( d \) by evaluator \(j\).
    \item \( m \) is the number of dimensions, 5 in our case.
    \item \( R_{i,j} \) is the rank of the counter-argument \( i \) calculated by summing \( T_{i,j, d} \)  from evaluator \( j \).
\end{itemize}


The counter-arguments are ranked in ascending order according to their calculated \( O_{i,j} \) values, where lower scores correspond to superior performance. This ordinal ranking methodology effectively normalizes individual scoring variations and minimizes evaluator bias, as it accounts for potential systematic differences in scoring tendencies among evaluators who might consistently assign either higher or lower scores, thereby ensuring a fair comparison.

The evaluation using reference-based metrics involves calculating scores for counter-arguments generated by each LLM in comparison to the gold reference, which enables the establishment of relative rankings among the LLMs' outputs.

\section{Results}\label{sec:results}

We first calculate the correlation of the automatic metrics with human judgments. We then used the best automatic metrics to compute the overall rankings for the gold counter-arguments and those generated by the four models. Finally, we analyzed dimension-wise rankings obtained in the manual evaluation.

\begin{figure}[ht]
        \centering
         \includegraphics[scale=0.55]{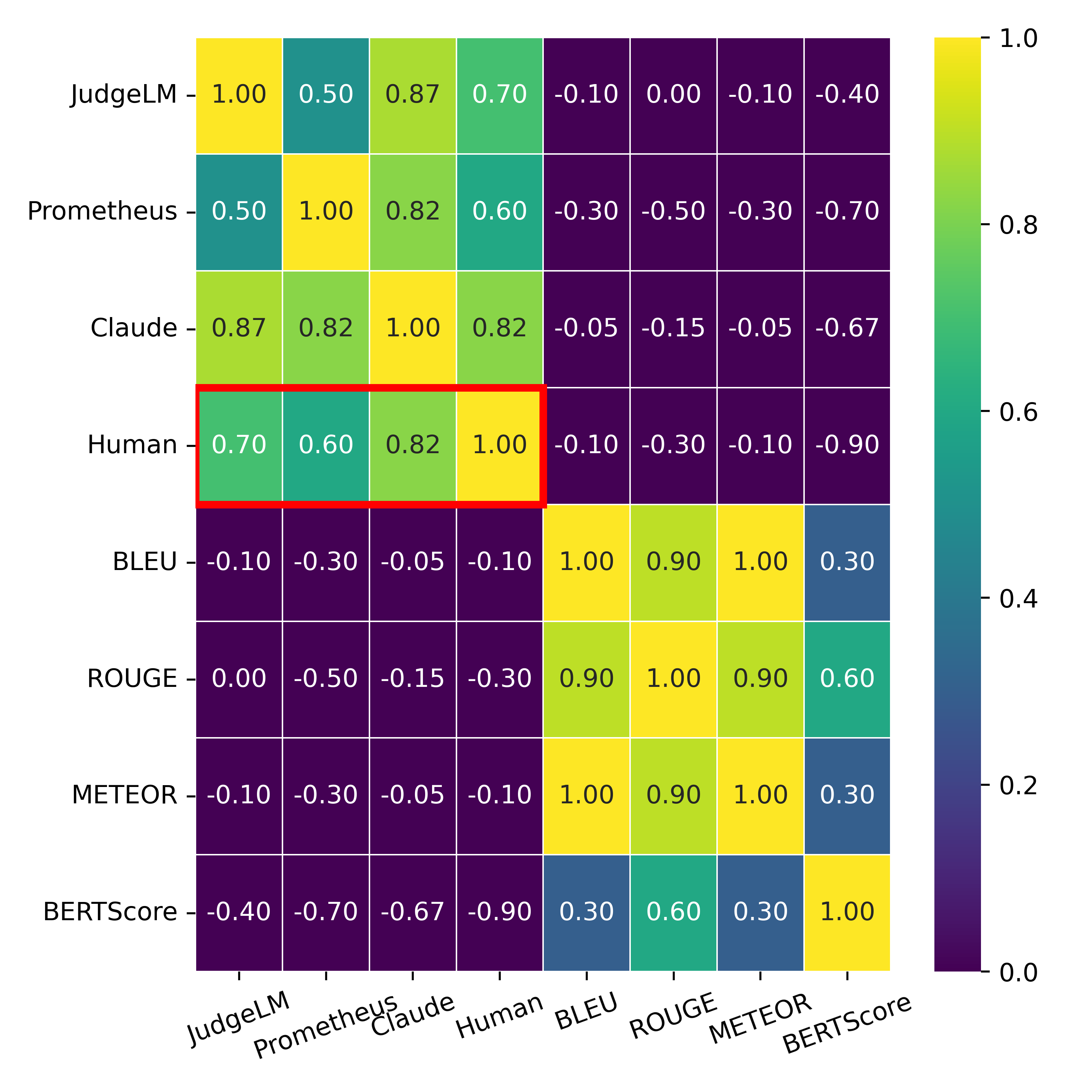}
        \caption{Heatmap showing Spearman's rank correlation coefficients between human evaluation and automatic metrics, including LLM-as-a-Judge metrics. The row marked in red represents the correlation of all the evaluation metrics to human preference.}
        \label{fig:human_llm_heatmap}
\end{figure}

\noindent \textbf{Correlation with Human Judgments.} We would like to stress that manual counter-argument evaluation is a highly subjective and tedious process \cite{wachsmuth-etal-2017-argumentation,hua-etal-2019-argument-generation,chen-etal-2024-exploring-potential}. Thus, ensuring that our automatic evaluation method correlates with human judgments is crucial. Therefore, we computed the correlation between every evaluation metric on the evaluator sample set of 75 examples, including human evaluation, using the method described in Section \ref{sec:experimental-setup}. The  Spearman's rank correlation coefficients ($\rho$) are reported in Figure \ref{fig:human_llm_heatmap}. First, it can be observed that the three LLM-as-a-Judge methods obtain the highest correlation with human judgments (row marked in red). Thus, while JudgeLM and Prometheus obtain a strong correlation, Claude 3.5 Sonnet is the best method with a $\rho$ of 0.82 (very strong correlation).

\begin{figure*}[ht]
\centering
    \includegraphics[scale=0.35]{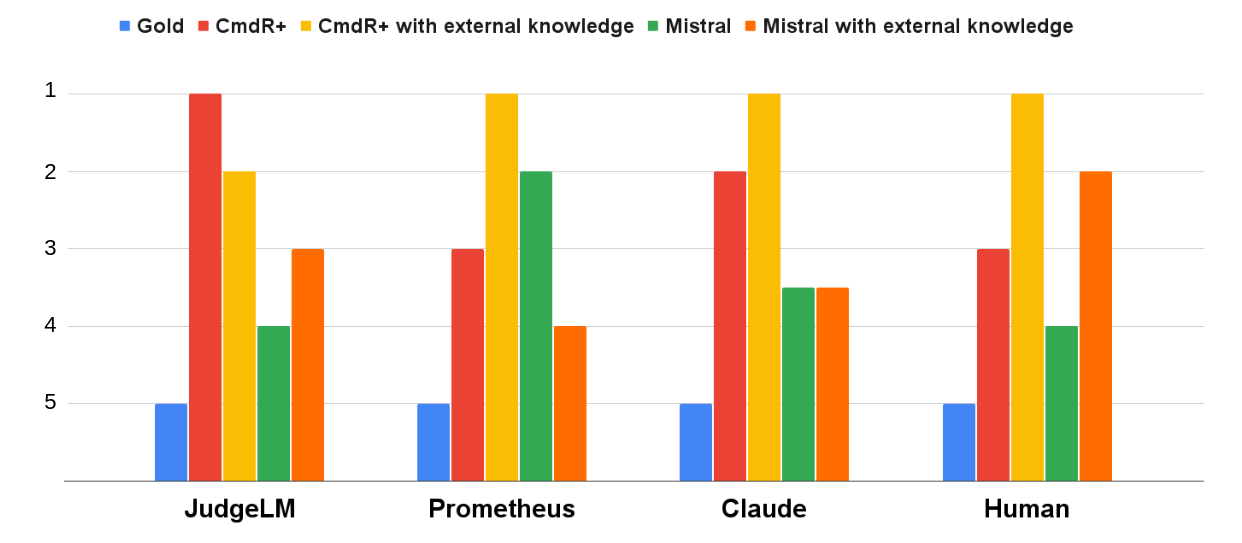}
    \caption{Human and LLM-as-a-Judge evaluation results.}
    \label{fig:ranks_llms_anns}
\end{figure*}

Reference-based metrics show a poor correlation with both human judgments and LLM-as-a-Judge evaluation methods. This suggests that reference-based metrics may be inadequate for evaluating the quality of counter-argument generation, as they fail to capture the essential dimensions established for human evaluation and disregard the context of the original argument or claim being countered.
 
Among the LLM-as-a-Judge methods, the lowest correlation was observed between JudgeLM and Prometheus. Upon a more detailed analysis, we found that these two judges had the highest level of disagreement in evaluating Opposition and Persuasiveness. Prometheus tends to be stricter on Opposition, while JudgeLM does the same on Persuasiveness. Thus, each model assigned lower scores than the other for those particular dimensions.

\noindent \textbf{Overall rank evaluation.} Figure \ref{fig:ranks_llms_anns} reports the overall rankings obtained by summing all the scores from each dimension that each evaluator provided (as described in Section \ref{sec:experimental-setup}). The first observation is that 2 out of 3 LLM-as-a-Judge agree that \commandrev generates the best counter-arguments. Moreover, according to the manual evaluation, both \commandr and \mistral with external knowledge are ranked at the top. Interestingly, results with respect to the ranking of other models vary, but all four evaluators agree that the human-generated counter-arguments (gold reference) are ranked worst.

Table \ref{tab:automatic_scores} shows the scores obtained by the reference-based metrics. While \commandrev remains the best model, it is easy to see that the rankings are not aligned with respect to human preferences. Furthermore, although not directly comparable, the obtained scores are in the range of previous work evaluating counter-argument generation with these metrics \cite{chen-etal-2024-exploring-potential}.


\begin{table*}[ht]
  \centering
  \begin{tabular}{l|cccc|c}
    \toprule
           Model    &  BLEU & ROUGE & METEOR & BERTScore & Avg \\
    \midrule
        \commandr &  20.35  & 18.36 & 16.12 & \underline{\textbf{86.38}} & 35.30 \\
        \commandrev   &  \underline{\textbf{20.80}}  & \underline{\textbf{18.67}} & \underline{\textbf{16.81}} & 86.15 & \underline{\textbf{35.60}} \\
        \mistral &  \underline{17.36}  & 15.93 & 13.96 & 86.23 & 33.37  \\
        \mistralev &  17.30  & \underline{16.58} & \underline{14.36} & \underline{86.29} & \underline{33.63} \\
    \midrule
  \end{tabular}
  \caption{Results with reference-based metrics; best overall model per metric in \textbf{bold}; best model per family \underline{underlined}.}
  \label{tab:automatic_scores}
\end{table*}


Taking into account the high correlation of LLM-as-a-Judge methods with human preferences, we also evaluated the four models over the full corpus of 150 arguments and counter-arguments pairs. The results from each LLM-as-a-Judge evaluator can be found in the Appendix \ref{sec:appendix} (Figures \ref{appendix:judgelm_per_dim_ranks}, \ref{appendix:prom_per_dim_ranks}, \ref{appendix:claude_per_dim_ranks}, \ref{appendix:all_overall_ranks}).
    
    \begin{figure}[ht]
        \includegraphics[width=1\linewidth]{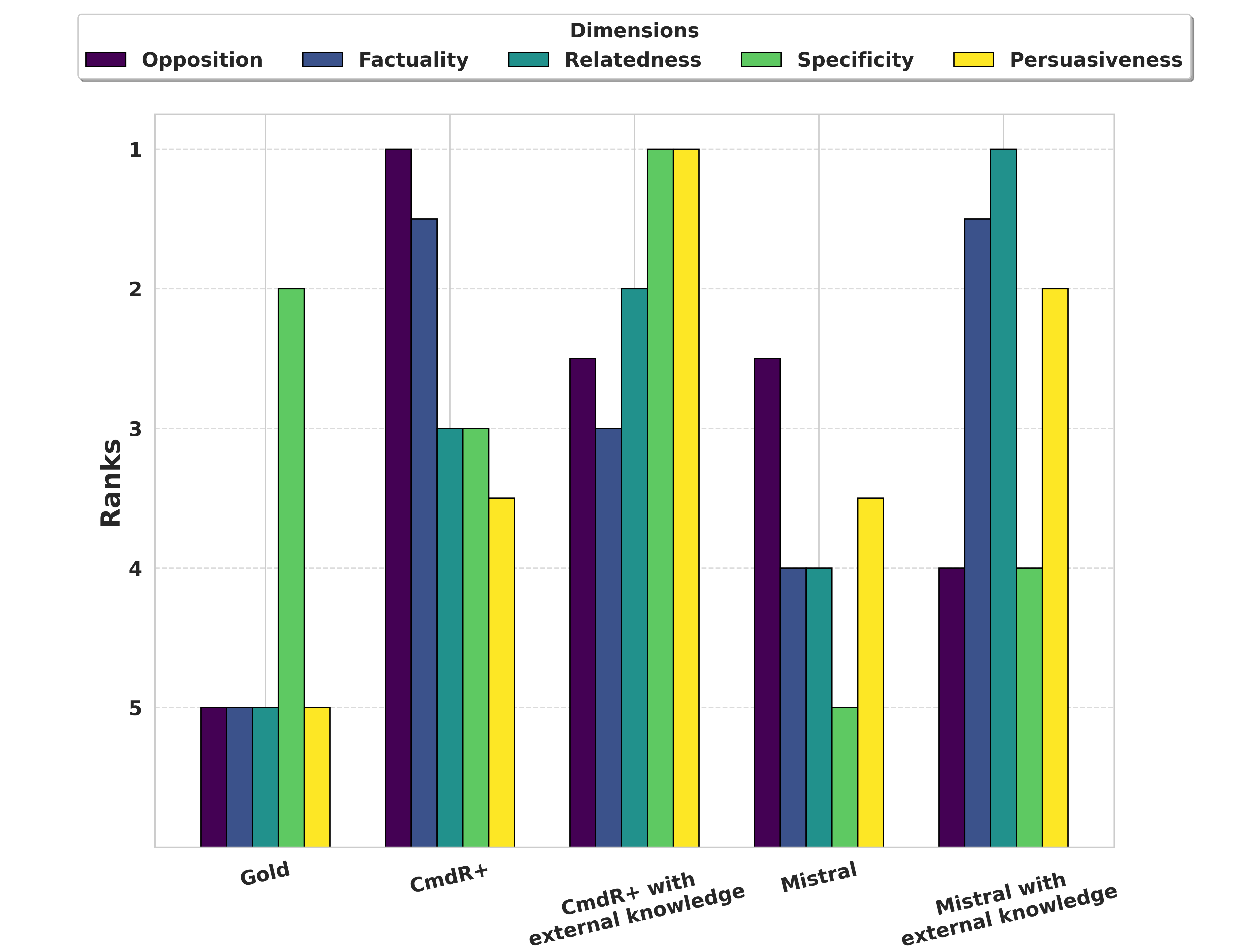}
        \centering
        \caption{Per dimension ranking from manual evaluation.}
        \label{fig:per_dim_comparison}
    \end{figure}

\noindent \textbf{Point-wise evaluation}. Figure \ref{fig:per_dim_comparison} illustrates the average dimension-wise rankings from manual evaluation. We can see that \commandr excels in Opposition compared to others, whereas both \commandr and \mistral with external knowledge are valued the best on Persuasiveness and Relatedness. Moreover, \mistralev and \commandr obtain the highest Factuality scores. Overall, we can conclude that when external evidence is provided to the LLMs, the overall quality of the counter-argument improves. Nevertheless, the performance of generating counter-arguments based on parametric knowledge is quite high, especially for the larger models (\commandr).


\section{Analysis of Results}

To answer RQ3, we checked whether the generated counter-arguments indeed used the provided external knowledge or whether the retrieved evidence is already incorporated in the parametric knowledge of LLMs. 


We used the sentence transformer model gte-base-en-v1.5' \cite{li2023gte, zhang2024mgte} to calculate sentence-level cosine similarity between the generated outputs and provided external evidence. Our methodology involved segmenting both the external information and generated counter-arguments into individual sentences for pairwise comparison, followed by ranking the similarity scores in descending order. Based on manual verification, we established a threshold of 70\% similarity as indicative of successful external knowledge integration. Our analysis revealed that \commandrev effectively utilized external evidence in 82\% of its generated counter-arguments, while \mistralev demonstrated such integration in 51\% of the cases.


Our manual analysis of similarity scores revealed that, when similarity scores exceed 70\%, models directly integrate provided external knowledge, while scores between 65-70\% result in partial incorporation mixed with additional information. Notably, this partial incorporation behavior predominantly occurs with sensitive topics such as personal reputation, religion, ethics, economics, or politics, where models tend to generalize rather than use specific factual evidence, leading to lower Opposition dimension scores. \commandrev frequently exhibited this behavior, though interestingly, the same model without external knowledge produced more generalized responses that evaluators often found more plausible than fact-based counterarguments.

The topics of high controversy trigger LLMs to start generation by acknowledging the stance of the input argument, which exhausts the space for including the retrieved evidence, even when explicitly prompted not to do so. Moreover, such counter-arguments were evaluated better than the output that starts the alternative perspective straight from the factual examples.  

\section{Conclusion}


In this study, we examined the effect of incorporating externally dynamically retrieved evidence into LLMs in counter-argument generation using web search as an external knowledge source. 
Our results on a newly created gold dataset show that, while LLMs with external knowledge improved their counter-argument generation, their reliance on it varies, and in some cases, they generate responses with parametric knowledge that obtained better scores.

We propose LLM-as-a-Judge to automatically evaluate the quality of counter-arguments with better correlation scores with respect to human judgments than previously used reference-based metrics.

Through qualitative analysis, we found that model behaviour shifts when dealing with sensitive or controversial topics. In these cases, LLMs tend to provide more generalized responses rather than directly integrating factual evidence. Interestingly, such responses were often rated more favourably, suggesting a preference for plausibility and coherence over strict factual accuracy.

Our findings highlight the complexities of integrating external knowledge into LLM outputs. While retrieval-augmented generation (RAG) can enhance factual consistency, models may still prioritize linguistic fluency and alignment with social norms. Future work should focus on refining strategies to ensure that external knowledge is utilized effectively, particularly in contexts that require precise and evidence-based argumentation.

\section{Limitations}

Our study has some limitations. We focused on two LLMs, both with and without integrated external knowledge, to compute the agreement between human and LLM judgments. Including more models would have significantly strengthened our conclusions. Equally, including other languages should allow for more generalizable results. However, as far as we know, there are no counter-argument generation datasets for languages beyond English. Therefore, one of the short-term objectives of the NLP research community should be to address this glaring gap.

Additionally, due to the effort required for manual assessment, we evaluated only a sample of the generated counter-arguments rather than the entire dataset. Future work could explore more scalable evaluation methods to extend the analysis.

Finally, the LLM-generated counter-arguments may be affected by potential data contamination, where topics and examples of the arguments used in our experiments may overlap with the training data of the LLMs we used. Investigating task contamination is far from trivial, but it should be included in any future work.

\section*{Acknowledgments}

We are grateful for the provided API credits by the Cohere For AI Research Grant Program\footnote{\url{https://cohere.com/research/grants}}.

We would also like to acknowledge the funding received by the following MCIN/AEI/10.13039/501100011033 projects: {(i) DeepKnowledge (PID2021-127777OB-C21) and ERDF A way of making Europe; (ii) DeepMinor (CNS2023-144375) and European Union NextGenerationEU/PRTR; (iii) Disargue (TED2021-130810B-C21) and European Union NextGenerationEU/ PRTR; (iv) EDHIA (PID2022-136522OB-C22); (v) LOTU (TED2021-130398B-C22) and European Union NextGenerationEU/ PRTR. Anar Yeginbergen's PhD contract is part of the PRE2022-105620 grant, financed by MCIN/AEI/10.13039/501100011033 and by the FSE+.

\bibliography{custom}

\clearpage

\appendix
\onecolumn 
\section{Example of data}
\label{appendix: data_example}

Table \ref{tab: data_example} shows the sequential methodology employed in generating the corpus. \textit{Original} presents an argument from the publicly available \texttt{CANDELA} corpus that originally is in lowercase and segmented in sentences and tokens. After its reconstruction to a human-readable format, in the \textit{Intermediate} step, the argument is summarized by means of the language model Llama-3.1-70B instruct. The row \textit{Final} presents the version after manual elaboration which is made available in this paper.

\begin{table*}[htbp]
\begin{footnotesize}
    \begin{tabular}{p{4cm} | p{11cm}} 
        \hline
        \textbf{Original} & 
        ["we", "should", "n't", "worry", "about", "being", "compassionate", "to", "mexican", "illegal", "immigrants",  
        "the", "same", "way", "we", "do", "n't", "worry", "about", "being", "uncompassionate", "to", "the", "rest",  
        "of", "the", "world", "'s", "poor", ".", "i", "am", "specifically", "referring", "to", "poor", "immigrants",  
        "who", ",", "based", "on", "current", "tax", "codes", ",", "will", "take", "far", "more", "in", "benefits",  
        "than", "they", "would", "pay", "in", "taxes", ".", "it", "has", "nothing", "to", "do", "with", "skin", "color",  
        "...", "if", "you", "have", "millions", "of", "white", "people", "suddenly", "all", "working", "manual",  
        "labor", "jobs", "and", "below", "you", "now", "have", "a", "lot", "of", "people", "not", "paying", "many",  
        "taxes", "into", "the", "system", "and", "being", "eligible", "to", "take", "a", "lot", "out", ".", "why",  
        "do", "people", "argue", "we", "need", "to", "be", "\"", "compassionate", "\"", "when", "with", "that",  
        "same", "logic", "you", "could", "argue", "we", "are", "n't", "being", "compassionate", "for", "not",  
        "all", "living", "a", "minimalist", "life", "and", "sending", "all", "our", "wealth", "to", "africa",  
        "until", "there", "are", "no", "more", "starving", "people", "?", "what", "makes", "Mexico", "so",  
        "deserving", "of", "our", "aid", "but", "not", "other", "countries", "?", "logically", "i", "'m",  
        "sure", "the", "people", "clamoring", "to", "he", "compassionate", "and", "let", "all", "the",  
        "poor", "immigrants", "in", "(", "i.e.", "making", "the", "immigration", "process", "easier",  
        "or", "amnesty", ")", "realize", "we", "could", "n't", "support", "the", "entire", "world",  
        ",", "so", "why", "is", "mexico", "special", "?", "what", "do", "you", "consider", "our",  
        "breaking", "point", "for", "percentage", "of", "us", "poor", "(", "i", "think", "we", "'re",  
        "already", "at", "it", ")", "to", "where", "there", "'s", "more", "money", "going", "out",  
        "to", "social", "programs", "and", "tax", "breaks", "than", "there", "is", "coming", "in",  
        "?", "at", "what", "point", "does", "the", "super", "pro", "immigration", "person",  
        "decide", "the", "us", "is", "\"", "full", "\"", "?"] 
        \\ 
        \hline
    \textbf{Intermediate}  & 
        The writer argues that showing compassion to Mexican illegal immigrants is not justified when considering the financial burden they would place on the system, as they would take more in benefits than they would pay in taxes. The sustainability of immigration policies is based on economic impact rather than emotions. Uncertainty about the threshold at which the U.S. would be considered "full" and unable to support more immigrants \\
        \hline
    \textbf{Final (Ours)} &
        Showing compassion to Mexican illegal immigrants is not justified when considering the financial burden they would place on the system, as they would take more in benefits than they would pay in taxes. This argument is not based on skin color, but rather on the economic impact of a large influx of low-income workers. \\
    \hline
        
    \end{tabular}
    \caption{Example of the data. \textit{Original} is the original publicly available data. \textit{Intermediate} is the summary generated by Llama-3.1-70B. \textit{Final} is the final version after manual refinement and the data used in the experiments.}
    \label{tab: data_example}
        \end{footnotesize}
\end{table*}

\section{Examples of arguments excluded due to subjectivity, irrelevance, or lack of verifiability}
\label{appendix: rejected_examples}
\begin{tcolorbox}[colback=white,colframe=black,arc=3mm, boxrule=0.8pt, width=\linewidth]
\begin{footnotesize}
\begin{itemize}
    \item if a logical outcome of an action or ideology is undesirable, people will try to find a way to avoid that outcome while still pursuing the action or ideology. They argue that people and societies are generally smart and reasonable enough to know when to stop before reaching an undesirable conclusion.
    \vspace{2mm}
    \item if Napoleon had won at Waterloo and completed his conquest of Europe, the continent would have flourished economically, socially, and politically.
    \vspace{2mm}
    \item Santa remains a beloved and benign figure in popular culture.
    
    \vspace{2mm}
    \item Eating meat is equivalent to murder, and therefore, meat-eaters feel hypocritical calling out others for immoral behavior. It is not fair to judge others for their moral transgressions, such as racism or rape, while they themselves contribute to animal suffering.

    \vspace{2mm}
    \item One should take ownership of their own problems and not place the burden on others to fix them.
    
\end{itemize}
\end{footnotesize}
\end{tcolorbox}

\section{Snippet of the final version of the dataset}
\label{appendix: more_data_examples}
An example of the final version of the data in Table \ref{tab:more_data_examples}. 

\begin{table*}[htbp]
\begin{footnotesize}
    \centering
    \begin{tabular}{p{7.5cm} | p{7.5cm}}
        \textbf{Argument} & \textbf{Counter-Argument} \\
        \hline
        \textcolor{blue}{[Increasing privilege for everyone is not possible without those who are currently benefiting losing some of their advantages]}, and that \textcolor{blue}{[people only pay lip service to values like equality and fairness without being willing to make real sacrifices.]} \textcolor{blue}{[When faced with the choice, most people will prioritize protecting their own privileged position rather than fighting for equality.]} & \textcolor{blue}{[Relinquishing some privilege can be in one's self-interest, as it can prevent violence and resentment from those who feel underprivileged.]} \textcolor{blue}{[Affirmative action and recognizing privilege can lead to a more equal society, where no one has advantages based on skin color, sexuality, or gender.]} \textcolor{blue}{[A society that promotes universal principles and equality can benefit everyone, including those in positions of power.]} \\
        \hline
         \textcolor{blue}{[Small countries in Europe should unite as a single nation, like the US]}, and \textcolor{blue}{[eventually all small countries should become large, united territories.]} &  \textcolor{orange}{[The US is a union of states with its own laws and cultures]}, but \textcolor{orange}{[it has a high homicide rate compared to the EU, partly due to clashes between different ethnic and socioeconomic groups.]} \textcolor{orange}{[Closer economic union between European countries may have advantages and disadvantages]}, but \textcolor{blue}{[a political union may not be beneficial as most countries feel they can govern themselves better than a central European government]}. \textcolor{blue}{[A separate union for smaller countries could be an option, providing financial leverage and allowing them to maintain their social programs and sovereignty.]} \\
        \hline
         \textcolor{blue}{[Food past recommended daily values should be heavily taxed]}, as \textcolor{orange}{[overeating is a major problem that leads to health issues and costs money in healthcare.]} \textcolor{red}{[I propose a system where individuals would be encouraged to eat the proper amount of food, as calculated by a doctor's visit]}, and \textcolor{red}{[those who choose to eat excessively would be taxed to support the system and fund welfare programs.]} &  \textcolor{blue}{[The current food distribution system is flawed because it's based on wealth, favouring the rich]}, and \textcolor{blue}{[a proposed system to tax food based on consumption would not solve the issue of global hunger]}, which \textcolor{orange}{[is caused by factors like war and poor distribution.]} \textcolor{orange}{[The world produces enough food to feed everyone]}, but \textcolor{orange}{[over 50\% of it is wasted]}, and \textcolor{orange}{[the problem of obesity is not solely caused by food consumption.]} Instead of taxing food, \textcolor{blue}{[it would be more effective to work on improving the distribution system]} and \textcolor{blue}{[addressing the root causes of global hunger.]} \\
        \hline
         \textcolor{blue}{[Capitalism is the best economic option]}, as \textcolor{blue}{[it is a free market regulated by the state]} that \textcolor{blue}{[allows individuals to become self-made millionaires.]} \textcolor{blue}{[A just and non-corrupt state is necessary for a successful capitalist environment.} \textcolor{blue}{Communism, on the other hand, interferes with individual freewill and kills incentives]}, while \textcolor{blue}{[a laissez-faire system is unattainable due to human nature.]} &  \textcolor{orange}{[Government intervention interferes with the free market]}, which \textcolor{blue}{[is contrary to pure capitalism.]} \textcolor{orange}{[Historically, capitalism has been the best option to incentivize the production of physical goods in a world with abundant natural resources.]} However, \textcolor{orange}{[in a future where physical work is performed by robots and software only needs to be written once]}, \textcolor{blue}{[capitalism may not be the best option.]} \\
    \end{tabular}
    \caption{Final argument and counter-argument pair that was obtained as described in Section \ref{sec:data}. Text in \textcolor{blue}{Blue} highlights the claim, evidence in \textcolor{orange}{Orange}, and examples are in \textcolor{red}{Red}. Criteria to identify claims and premises based on \citet{sviridova-etal-2024-casimedicos}.}
    \label{tab:more_data_examples}
\end{footnotesize}
\end{table*}

\FloatBarrier
\newpage
\section{Prompts used for query generation, web search, generation with and without external knowledge}
\label{appendix: prompt_examples}

Table \ref{tab:prompt_list} illustrates the prompts used for each setting of the generation step.  

\begin{table*}[htbp]
\begin{footnotesize}
    
    \centering
    \begin{tabular}{p{4cm} | p{11cm}}
        \textbf{Setting} & \textbf{Prompt} \\
        \hline
        Query Generation & Generate a list of 5 queries for web-search that would help to find information to question the veracity of the given claim and persuade to take the opposing position: \{claim\}? \newline
                                Provide only questions and nothing else.
                                The answer should be in JSON. \\
        \hline
        Web-search & Answer the question from the following text: \{question\}? \newline
                                1. Find factual information from different media outlets \newline
                                2. Provide the evidence in a bullet point manner. \newline
                                3. Do not output anything else. \newline \\
        \hline
        Generation with external knowledge & Given the following context: \{qacontext\}, provide a succinct counter-argument that refutes the following argument using information from the context: \{claim\}. \newline
                                   Provide only the answer and nothing else. \newline
                                   Make sure the answer is no longer than 3 sentences. \\
        \hline
        Generation without external knowledge & Generate a succinct counter-argument that refutes the following claim: \{claim\}? \newline
                                   Provide only the answer and nothing else. \newline
                                   Make sure the answer is no longer than 3 sentences.\\
    \end{tabular}
    \caption{List of prompts used at every stage of the generation. \{claim\} is a placeholder for the input claim. \{question\} is a placeholder for the question. \{qacontext\} is a placeholder for the retrieved context.}
    \label{tab:prompt_list}
\end{footnotesize}
\end{table*}
\FloatBarrier
\section{Per dimension evaluation ranks}
\label{sec:appendix}

In this appendix, we provide evaluation ranks per opposition, factuality, relatedness, specificity, and persuasiveness dimensions for the following LLM-as-a-Judge models: JudgeLM in Figure \ref{appendix:judgelm_per_dim_ranks},   Prometheus in Figure \ref{appendix:prom_per_dim_ranks}, and Claude in Figure \ref{appendix:claude_per_dim_ranks}. The overall ranks over the whole dataset by LLM-as-a-Judge are shown in Figure \ref {appendix:all_overall_ranks}.

Results, especially those computed using Claude 3.5 Sonnet, align with those obtained in Section \ref{sec:results} with a human-annotated sample of the data. Thus, \ref {appendix:all_overall_ranks} shows that \commandrev is also ranked first, gold reference, and \mistral the worst, while the ranks for \commandr and \mistralev vary according to the judge.

\begin{figure}[h]
        \includegraphics[width=0.55\linewidth]{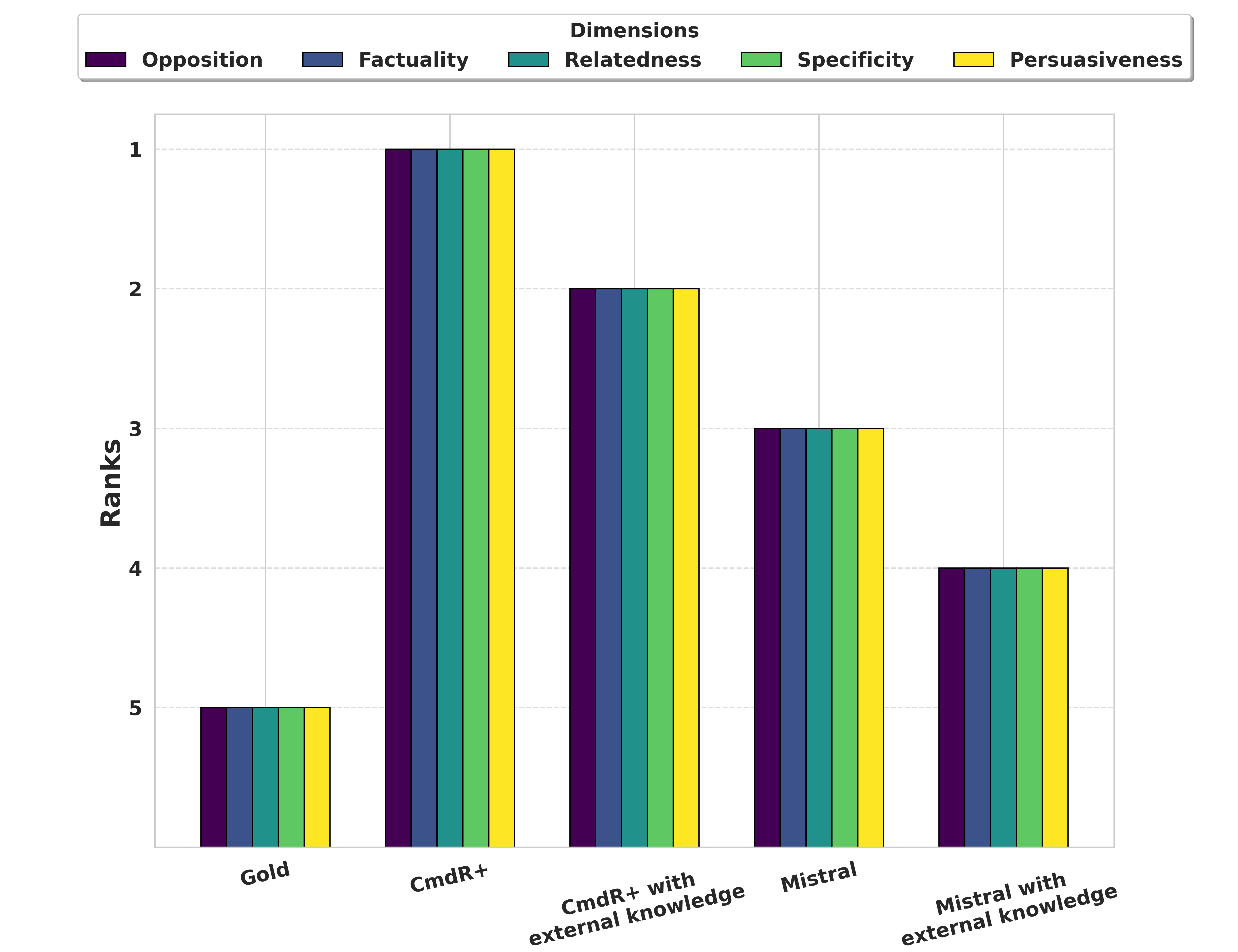}
        \centering
        \caption{JudgeLM per dimension ranks.}
        \label{appendix:judgelm_per_dim_ranks}
\end{figure}

\begin{figure}[h]
        \includegraphics[width=0.55\linewidth]{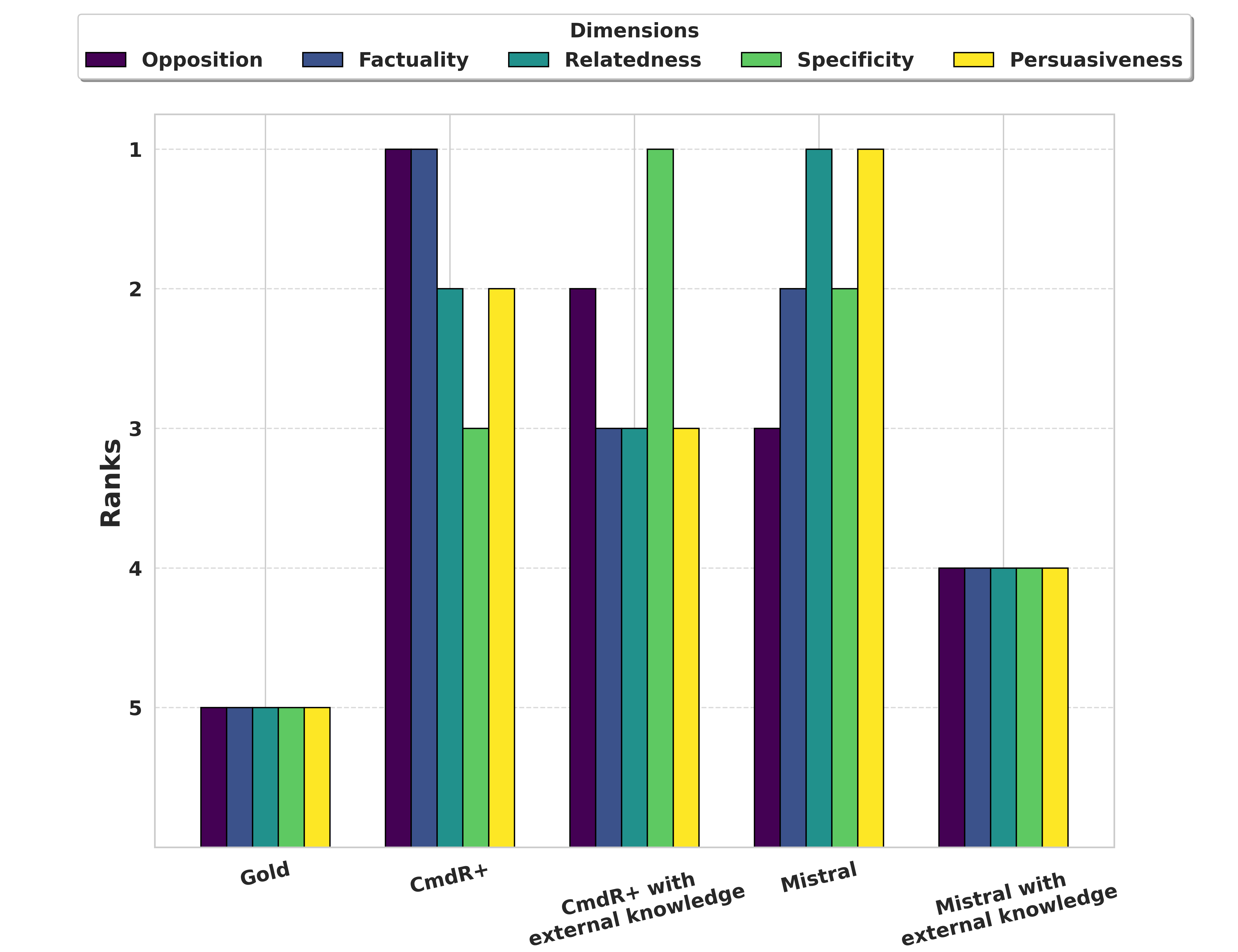}
        \centering
        \caption{Prometheus per dimension ranks.}
        \label{appendix:prom_per_dim_ranks}
\end{figure}

\begin{figure}[h]
        \includegraphics[width=0.55\linewidth]{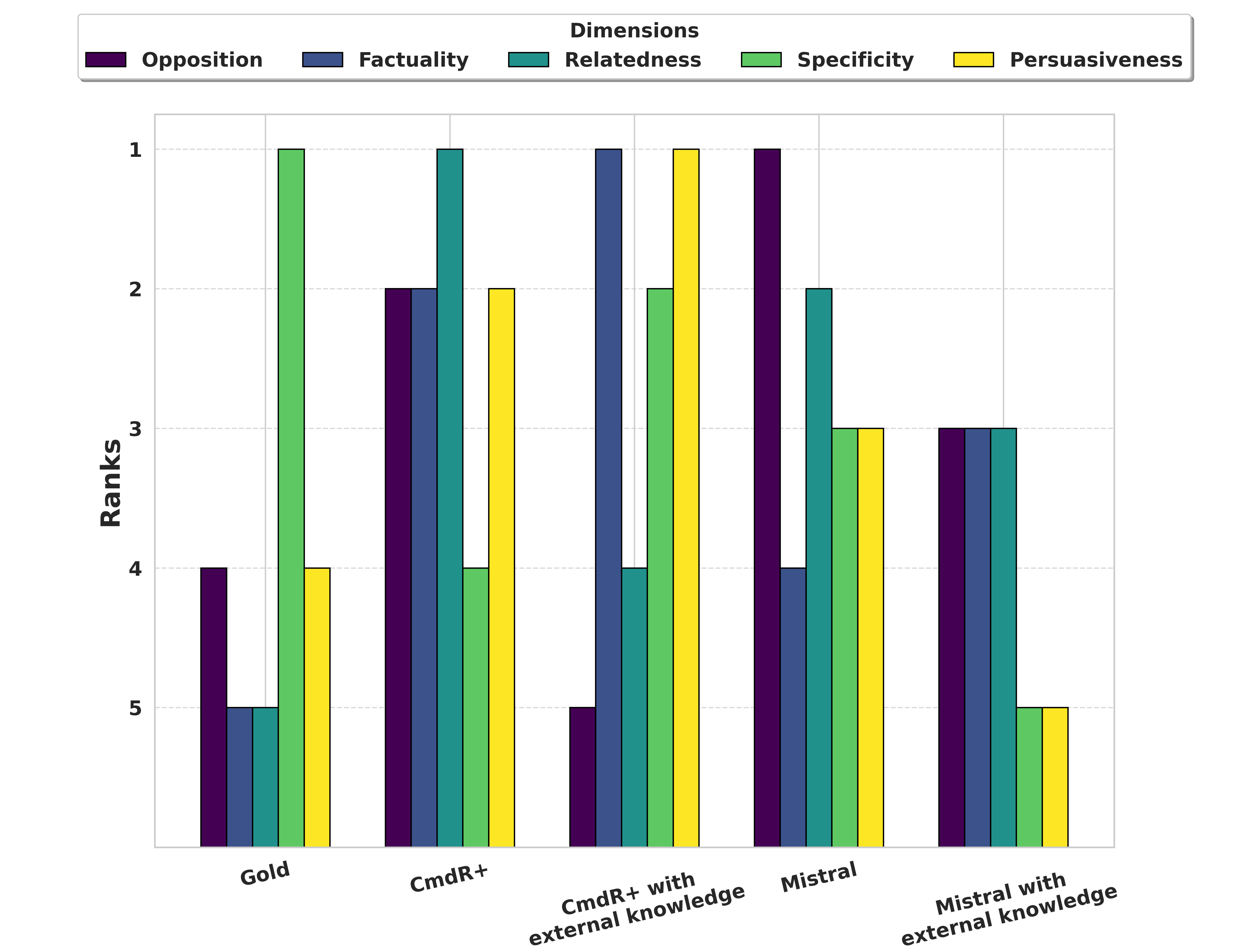}
        \centering
        \caption{Claude per dimension ranks.}
        \label{appendix:claude_per_dim_ranks}
\end{figure}
\onecolumn
\begin{figure*}[h!]
        \includegraphics[width=0.75\linewidth]{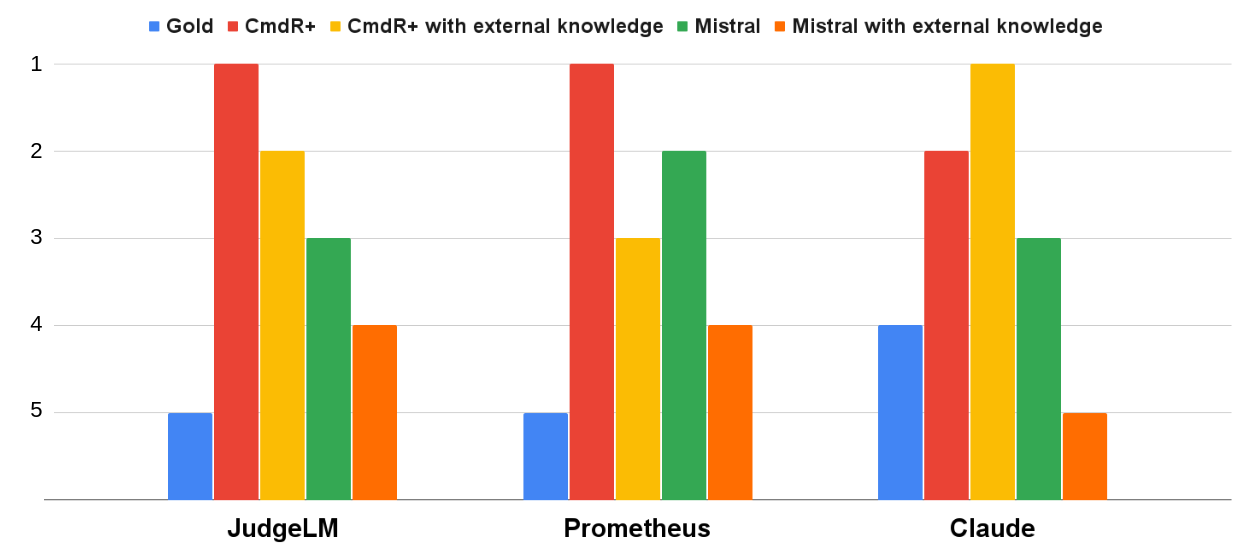}
        \centering
        \caption{LLM-as-a-Judge evaluation results over the full dataset.}
        \label{appendix:all_overall_ranks}
\end{figure*}

\FloatBarrier
\onecolumn

\section{Instruction example for human evaluation.}
\label{appendix: google_form_example}

An example of the instructions provided to the human evaluators in the Prolific platform is shown in Figure \ref{fig:instructions}. Firstly, the argument is presented accompanied by a description of the five dimensions (opposition, relatedness, factuality, specificity, persuasiveness) that are evaluated in the paper. Secondly, each of the counter-arguments is presented and the evaluator has to select a value between 1 for \textit{unsatisfactory}, 2 for \textit{moderately satisfactory}, and 3 for \textit{highly satisfactory} in each of the dimensions.

\begin{figure*}[h]
      \includegraphics[width=0.75\linewidth]{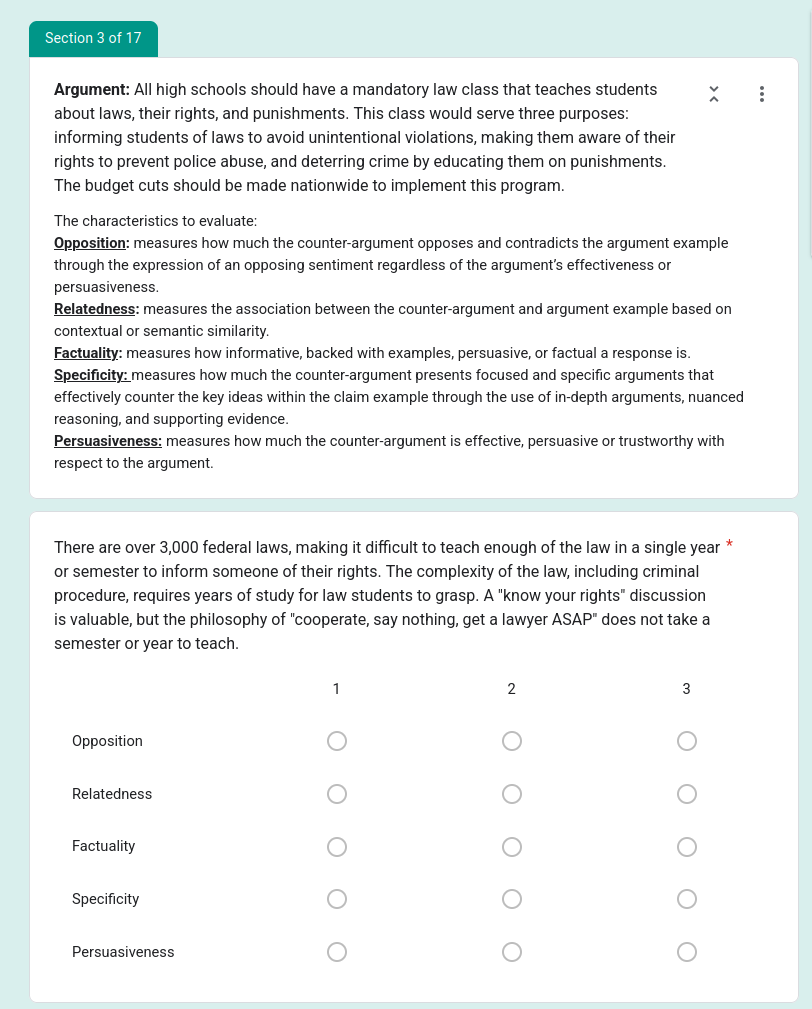}
        \centering
        \caption{An example of instructions provided to the human evaluators.}
         \label{fig:instructions}
\end{figure*}

\end{document}